\documentclass[letterpaper, 10 pt, conference]{ieeeconf}  

\IEEEoverridecommandlockouts                              

\usepackage{blindtext}
\usepackage{pdfpages}

\usepackage{graphics} 
\usepackage{graphicx}
\usepackage{lipsum} 
\usepackage{xcolor}
\usepackage{float}

\usepackage{amsmath,amsthm,amssymb}

\usepackage{hyperref}
\usepackage{cleveref}
\usepackage{cite}
\makeatletter
\def\@citex[#1]#2{\leavevmode
\let\@citea\@empty
\@cite{\@for\@citeb:=#2\do
{\@citea\def\@citea{,\penalty\@m\ }%
\edef\@citeb{\expandafter\@firstofone\@citeb\@empty}%
\if@filesw\immediate\write\@auxout{\string\citation{\@citeb}}\fi
\@ifundefined{b@\@citeb}{\hbox{\reset@font\bfseries ?}%
\G@refundefinedtrue
\@latex@warning
{Citation `\@citeb' on page \thepage \space undefined}}%
{\@cite@ofmt{\csname b@\@citeb\endcsname}}}}{#1}}
\makeatother

\title{\LARGE \bf
KEMP: Keyframe-Based Hierarchical End-to-End Deep Model for Long-Term Trajectory Prediction 
}

\author{Qiujing Lu$^{*,1,2,	\dagger}$, Weiqiao Han$^{*,1,3,	\dagger}$, Jeffrey Ling$^{1}$, Minfa Wang$^{1}$, Haoyu Chen$^{1}$,\\ Balakrishnan Varadarajan$^{1}$, Paul Covington$^{1}$ \\
$^{1}$Waymo, $^{2}$UCLA, $^{3}$MIT
\thanks{* Equal contribution}
\thanks{$\dagger$ Work done during internship at Waymo. Corresponding to \href{mailto:weiqiaoh@mit.edu}{weiqiaoh@mit.edu}, \href{mailto:qiujing@ucla.edu}{qiujiing@ucla.edu}}
\thanks{This article solely reflects the opinions and conclusions of its authors and not Waymo or any other Waymo entity.}
}

\begin{document}

\maketitle
\thispagestyle{empty}
\pagestyle{empty}

\begin{abstract}
Predicting future trajectories of road agents is a critical task for autonomous driving.
Recent goal-based trajectory prediction methods, such as DenseTNT and PECNet \cite{gu2021densetnt,mangalam2020not}, have shown good performance on prediction tasks on public datasets. However, they usually require complicated goal-selection algorithms and optimization.
In this work, we propose KEMP, a hierarchical end-to-end deep learning framework for trajectory prediction. At the core of our framework is \textit{keyframe-based trajectory prediction}, where keyframes are representative states that trace out the general direction of the trajectory.
KEMP first predicts keyframes conditioned on the road context, and then fills in intermediate states conditioned on the keyframes and the road context.
Under our general framework, goal-conditioned methods are special cases in which the number of keyframes equal to one.
Unlike goal-conditioned methods, our keyframe predictor is learned automatically and does not require hand-crafted goal-selection algorithms.
We evaluate our model on public benchmarks and our model ranked 1st on Waymo Open Motion Dataset Leaderboard (as of September 1, 2021).  

\end{abstract}

\section{INTRODUCTION}
In order for robots to navigate safely in stochastic environments with multiple surrounding moving agents, predicting future trajectories of surrounding agents is a critical task. 
In the setting of autonomous driving, the road scene is highly complex, consisting of not only static objects, such as traffic lights and road fences, but also dynamic objects, such as vehicles, pedestrians and cyclists; any vehicle could choose to go straight and pass the intersection, or stop before the intersection and wait for the pedestrians to pass, or make turns. 
Predicting future trajectories of agents in the scene enables several downstream tasks, such as risk assessment of planned trajectories \cite{wang2020fast} and safe trajectory planning for autonomous vehicles with theoretical guarantees \cite{lew2020chance,jasour2021convex}. 

Due to the dynamic, stochastic, and interactive nature of the environment, predicting future trajectories of agents based on past observations and the traffic scene is quite challenging. 
Traditional methods use hand-crafted features and manually-designed logic and models to predict trajectories \cite{deo2018would,yamaguchi2011you,ma2017forecasting}, but they require a great deal of manual work and are brittle to edge cases.
On the other hand, modern deep learning methods have successfully demonstrated the ability to scale with larger datasets, including work on graph neural networks \cite{zeng2021lanercnn}, long short-term memory (LSTM) \cite{alahi2016social}, generative adversarial networks (GAN) \cite{gupta2018social}, variational autoencoders (VAE) \cite{lee2017desire, yuan2019diverse}, flows \cite{rhinehart2019precog,rhinehart2018r2p2}, or transformers \cite{ngiam2021scene} to predict trajectories.

\begin{figure}
    \centering
    \includegraphics[width=0.45 \textwidth ]{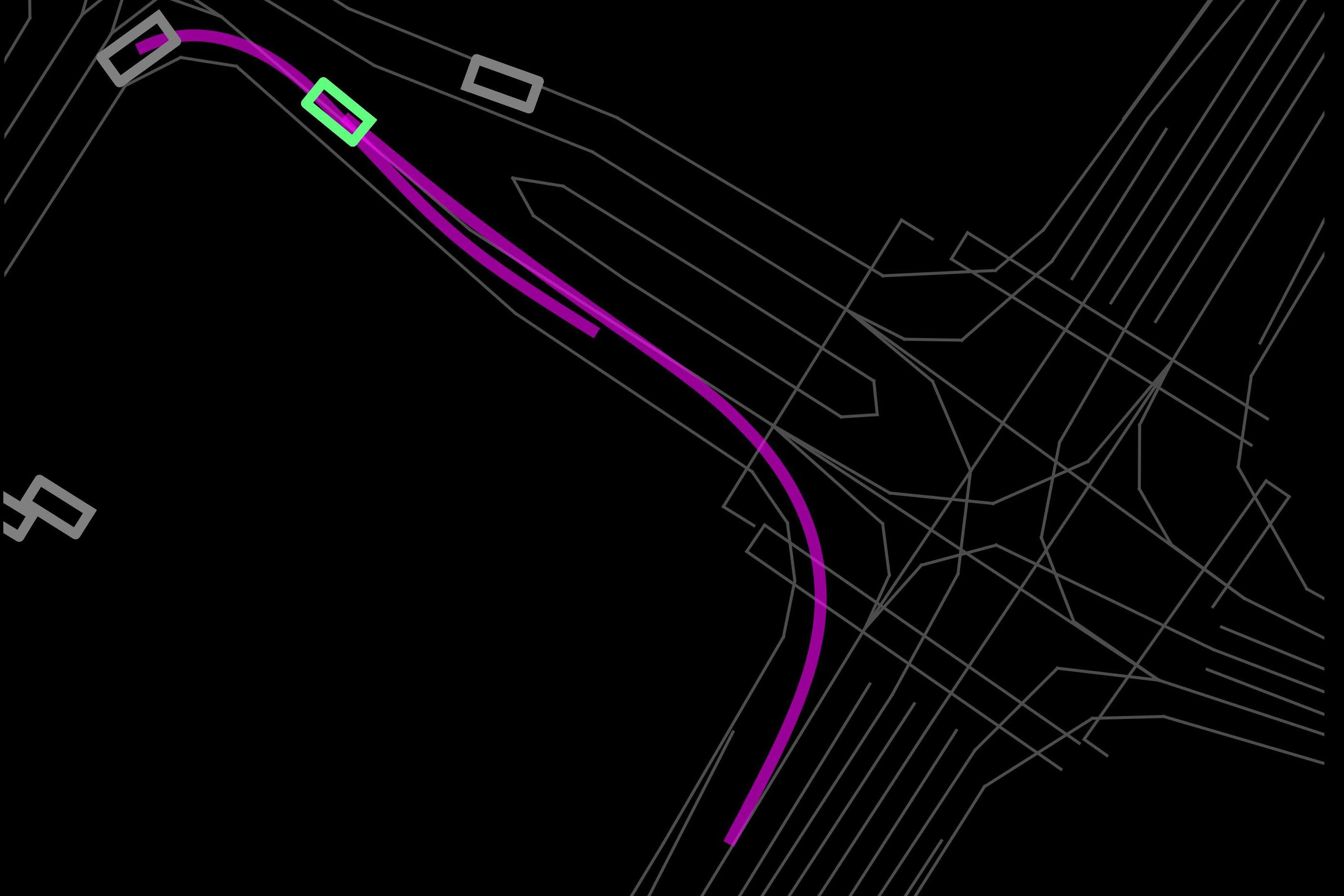}
    
    \vspace{0.1cm}
    \includegraphics[width=0.45 \textwidth ]{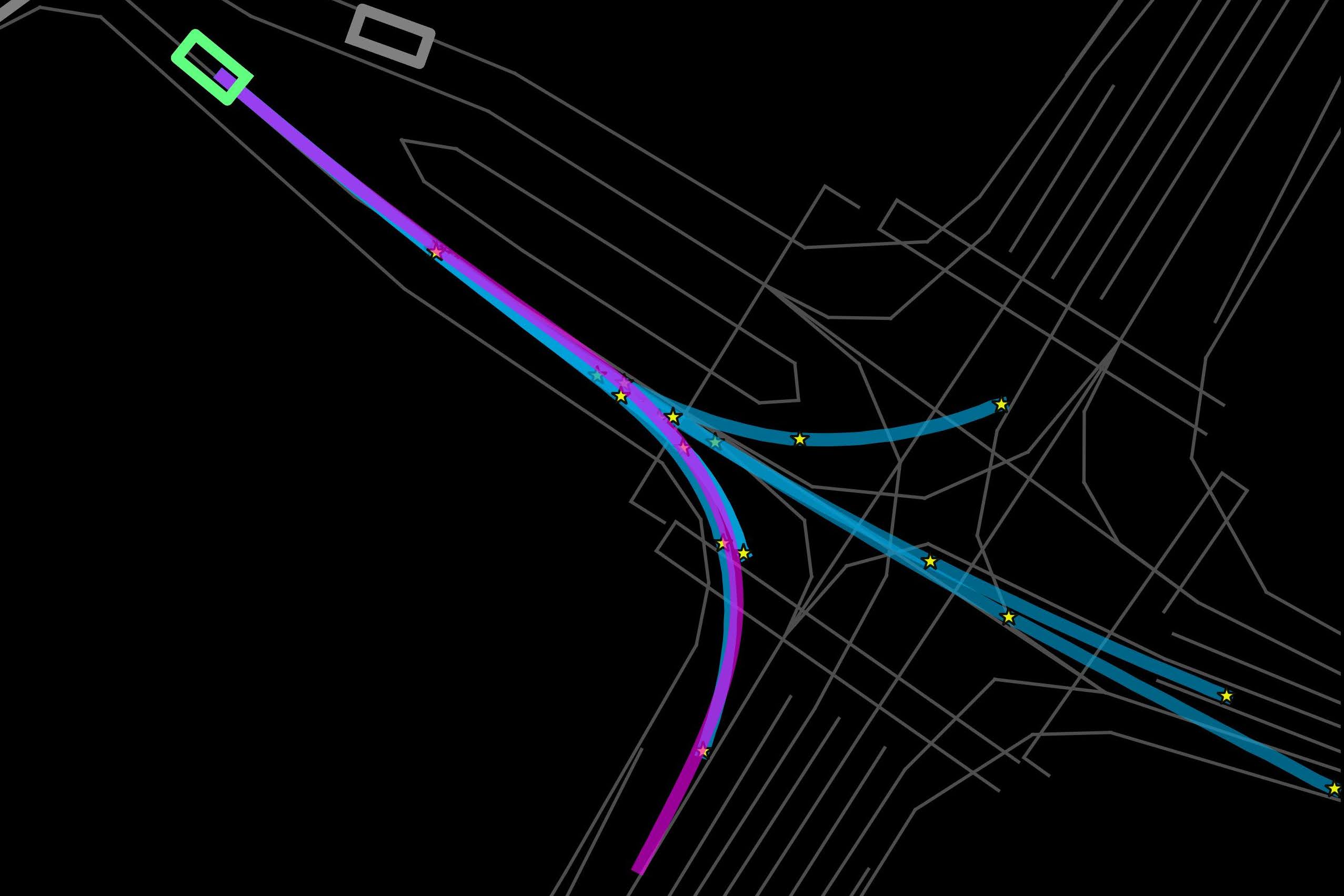}
    \caption{Agents in an intersection scenario. Top: Agents with ground truth future trajectories colored in magenta and the agent for the prediction task represented by a cyan box. Bottom: $6$ predicted trajectories colored in blue and keyframes annotated with yellow stars.}
    \label{fig:my_label}
\end{figure}
Recently, anchor-based and goal-conditioned methods \cite{chai2019multipath,zhao2020tnt,gu2021densetnt,rhinehart2019precog} have received much attention as they directly consider the intention of agents and are more interpretable.
However, when making long-term predictions (for example, in Waymo Open Motion Dataset \cite{ettinger2021large}, where one needs to predict 8 seconds into the future based on 1 second of past trajectories), only modeling a single high-level goal or intent may not be enough. For one thing, the goal prediction for long trajectories may not be accurate, and for another, trajectories can vary significantly between the fixed starting and goal points. 
To address this problem, we draw ideas from long-term motion planning and hierarchical reinforcement learning literature \cite{wang2019learning,nair2019hierarchical,kurutach2018learning,nachum2018data}, where in order for the robot to reach a goal far away or accomplish a complex task, a high level model generates subgoals that are easier for the robot to reach, and a low level model generates control inputs that enable the robot to navigate between two subgoals.

In this paper, we propose a hierarchical end-to-end deep learning framework for autonomous driving trajectory prediction: Keyframe MultiPath (KEMP). 
At the core of our framework is the keyframe-based trajectory prediction. 
In this framework, the model first predicts several \textit{keyframes}, which are representative states in the trajectory that trace out the general direction of the trajectory, conditioned on the road context. 
The model then fills in the gaps between keyframes by predicting intermediate states conditioned on the keyframes and the road context. 
To our best knowledge, it is the first time that keyframe-based hierarchical prediction is applied to trajectory prediction for autonomous vehicles. 
Our framework is in some sense a generalization of goal-conditioned trajectory prediction models. 
In particular, goal-conditioned trajectory prediction models, such as TNT \cite{zhao2020tnt}, DenseTNT \cite{gu2021densetnt}, and PECNet \cite{mangalam2020not}, can be viewed as special cases of keyframe-based trajectory prediction models where the number of keyframes equals to 1, but unlike these models, we allow the model to learn to predict keyframes instead of manually selecting goals.
Other trajectory prediction models that predict trajectories in one shot without conditioning on the final goal can be viewed as special cases of keyframe-based trajectory prediction models where the number of keyframes equals to 0. Our model is not only \textit{more general} than previous methods but also \textit{simpler} as keyframe prediction is learned automatically. 
Finally, our model achieves \textit{state-of-the-art performance} in autonomous driving trajectory prediction tasks, ranking 1st on Waymo Open Dataset Motion Prediction Leaderboard (as of September 1, 2021). 

\section{RELATED WORK}
\textbf{Latent-variable-sampling-based trajectory prediction}. A popular approach for trajectory prediction is sampling from latent variables.
DESIRE \cite{lee2017desire} generates trajectory samples via a conditional VAE-based RNN encoder-decoder.
R2P2 \cite{rhinehart2018r2p2} and PRECOG \cite{rhinehart2019precog} use flows to predict agent futures.
SocialGAN \cite{gupta2018social} uses recurrent generative adversarial networks to predict future trajectories.
These methods require stochastic sampling from latent distributions to produce implicit trajectories. The latent variables are not fully interpretable and hence do not work in combination with external prior knowledge. 

\textbf{Intention-based trajectory prediction}.
IntentNet \cite{casas2018intentnet} predicts intentions of drivers to guide trajectory prediction. They classify intentions into 8 classes, including keep lane, turn left, turn right, and so on. The method requires a great deal of manual engineering and might miss special cases on large datasets.
Multipath \cite{chai2019multipath} first predicts intents as a set of anchors and then fix the anchors and learn to predict the residual with respect to the anchors. 

\textbf{Goal-conditioned trajectory prediction}.
Goal-conditioned trajectory prediction models are a promising way to develop interpretable autonomous vehicle systems. PECNet \cite{mangalam2020not} predicts the goal as a latent variable and predicts the trajectory conditioned on this latent variable. TNT \cite{zhao2020tnt} and DenseTNT \cite{gu2021densetnt} predict a set of targets directly and then predict trajectories conditioned on the targets. Compared to latent-variable-sampling-based methods and intention-based methods, goal-conditioned methods such as are more interpretable, because the predicted goal is part of the trajectory instead of a latent variable. 
Our method can be viewed as a generalization of this line of work, where we predict not only the goal but also other keyframes in the trajectory. 
Unlike DenseTNT, in which there is a complicated goal-selection algorithm, our method automatically learns to predict keyframes without any hand-crafted engineering. 
\begin{figure*}
    \centering
    \includegraphics[width=0.98\textwidth]{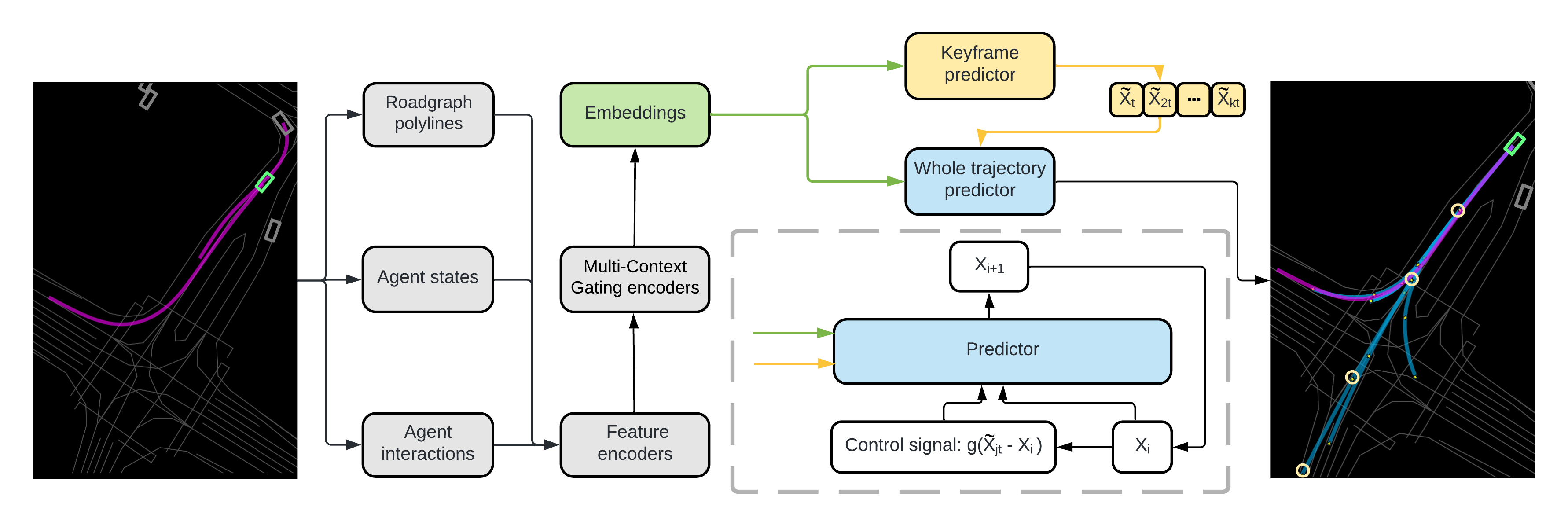}
   \caption{KEMP architecture. Context features are extracted from scenario inputs with multiple agent historical tracks by multiple encoders. They are then sent to our hierarchical decoders for generating predicted trajectories. The decoder consists of two parts: the keyframe decoder for the generation of keyframe locations and whole trajectory decoder for producing the final whole trajectory based on the previously decoded keyframe locations and context embeddings. In the predictor, we can feed in a control signal $g(\tilde{X}_{jt} - X_i)$ as a function of the distance to the subgoal.}
    \label{fig:arch}
\end{figure*}
\section{METHOD}
Our method consists of three steps. 
First, we extract the features from the scene and encode them as context using multiple encoders.
Second, we predict keyframes of the output trajectory using a keyframe predictor conditioned on the context.
Third, we predict intermediate states conditioned on keyframes and the context using the whole trajectory predictor. 
The whole model is trained end-to-end.

\subsection{Context Encoding}
To encode the road context, previous work uses rasterized encoding methods \cite{bansal2018chauffeurnet, chai2019multipath, chou2020predicting, cui2019multimodal, hong2019rules}. This method renders trajectories of moving agents and road context information as birds-eye view images and encodes them with CNNs.
Recently, vector-based representations, which represent the road and agents as polylines, has been more effective in capturing the structural features of high-definition maps \cite{gao2020vectornet,zhao2020tnt,gu2021densetnt}.
We adopt this vector-based sparse encoding representation.
Our context encoding mostly follows the methods in Multipath++ \cite{varadarajan2021multipath++}.

We use a deep neural network consisting of multiple copies of multi-layer perceptrons (MLPs) and max pooling layers to extract geometric features from road polylines and their connections with each agent.  
We use PointNet \cite{qi2017pointnet,qi2017pointnet++} to encode features from 2D points around each agent.
Each agent's raw state, including past positions, velocity, and heading, is encoded using an MLP. Interactions between agents are captured by encoding relative positions and speeds between pairs of agents using MLP and max pooling. 
All these features are mixed by going through several Multi-Context Gating (MCG) encoders, an efficient mechanism for fusing information \cite{varadarajan2021multipath++}.
In the end we concatenate all outputs from MCG encoders and get the context embedding $\mathbf{c}$.

\subsection{Keyframe-Based Hierarchical Trajectory Prediction}
In the prediction part, given the context $\mathbf{c}$, the goal is to predict $N$ trajectories $\ell_1,\ldots,\ell_N$. 
We follow the formulation in MultiPath \cite{chai2019multipath}. Each trajectory is the union of $T$ states $\ell_i = \{X_1,\ldots, X_T \}$, and each state $X_i$ is the tuple $(\mu_i, \Sigma_i)$, where $\mu_i$ is the expectation of the $(x,y)$ position of the agent at time $i$, and $\Sigma_i$ is the covariance matrix of the position prediction at time $i$. 

In our proposed method, the keyframe predictor predicts several \textit{keyframes}, which are defined as representative states in the trajectory that trace out the general direction of the trajectory, conditioned on the context $\mathbf{c}$. 
In this paper we focus on \textit{evenly spaced keyframes}.
More precisely, suppose $T = kt$, where $T$ is the total number of time steps for the prediction task and $k,t$ are two positive integers.
Then the keyframe predictor predicts $k$ keyframes $\tilde{X}_{t},\tilde{X}_{2t},\ldots,\tilde{X}_{kt}$ conditioned on the context $\mathbf{c}$.

We model the keyframes using a joint distribution
\begin{align*}
    \tilde{X}_{t},\tilde{X}_{2t},\ldots,\tilde{X}_{kt} \sim p(x_{t},x_{2t},\ldots,x_{kt}|\mathbf{c}).
\end{align*}
We can either use an autoregressive formulation
\begin{align*}
    \tilde{X}_{(i+1)t} \sim p(x_{(i+1)t} | \mathbf{c}, x_{t}, x_{2t}, \ldots, x_{it}), i=0,\ldots,k-1.
\end{align*}
or assume conditional independence between the keyframes
\begin{align*}
    \tilde{X}_{(i+1)t} \sim p(x_{(i+1)t} | \mathbf{c}), i=0, \ldots, k-1.
\end{align*}
In the former, an autoregressive predictor can be implemented with an LSTM, and in the latter, a non-autoregressive predictor can be implemented with a single MLP over all time steps.

Given the $k$ keyframes $\tilde{X}_{t},\tilde{X}_{2t},\ldots,\tilde{X}_{kt}$, we consider two ways to generate final trajectories.

\subsubsection{Interpolation Model}
For any interval $[\tilde{X}_{it}, \tilde{X}_{(i+1)t}]$, the whole trajectory predictor predicts the states inside the interval $X_{it+1}, \ldots, X_{(i+1)t-1}$ conditioned on $\tilde{X}_{it}$ and $\tilde{X}_{(i+1)t}$, as well as the context $\mathbf{c}$.
This gives us a complete trajectory 
\begin{align*}
    X_1, \ldots, X_{t-1}, \tilde{X}_{t}, X_{t+1},\ldots, X_{2t-1}, \tilde{X}_{2t}, X_{2t+1},\ldots, \tilde{X}_{kt}.
\end{align*}
The predictors predict $N$ trajectories $\ell_1,\ldots,\ell_N$. 
We assign a probability to each trajectory $p_i = p(\ell_i|\mathbf{c}) = {\exp f(\ell_i| \mathbf{c}) \over \sum_j \exp f(\ell_j|\mathbf{c})}$, where $f(\ell|\mathbf{c})$ is implemented by a deep neural network. 
Therefore, our prediction is a mixture of Gaussian distribution. 
We impose the negative log-likelihood loss on the predicted trajectory
\begin{align*}
    L_{traj}(\theta) = - \sum_{j=1}^N I(j=r) [\log p(\ell_i|\mathbf{c};\theta) + \\
    \sum_{i=1}^T \log \mathcal{N}(\bar{\mu}_i |\mu_i, \Sigma_i;\theta) ],
\end{align*}
where $\{\bar{\mu}_1,\ldots,\bar{\mu}_T\}$ represents the ground truth trajectory, $\theta$ represents the parameter to be learned, which is all the weights inside predictor models implemented by deep neural networks, including the whole trajectory predictor and the probability predictor. 
$r$ denotes the index of the trajectory that is closest to the ground truth measured by the $\ell_2$ distance. 

\subsubsection{Separable Model}
In the interpolation model, the keyframes in the final trajectory are predicted by the keyframe predictor. The whole trajectory predictor does not predict keyframes.  In the separable model, the whole trajectory predictor predicts the intermediate states $X_{it+1}, \ldots, X_{(i+1)t}$, including the keyframes, for any interval $[\tilde{X}_{it}, \tilde{X}_{(i+1)t}]$. This gives us a complete trajectory $X_1,\ldots,X_T$ predicted by the whole trajectory predictor. As an aside, when generating intermediate states, the whole trajectory predictor could condition on, in addition to the keyframes, some other manually defined control signals, such as a function of the distance to the subgoal $g(\tilde{X}_{jt} - X_i)$; in practice we use $g$ as the identity function.

As in the interpolation model, we impose the negative log-likelihood loss $L_{traj}(\theta)$ on the trajectories predicted by the whole trajectory predictor.
Different from the interpolation model, we also impose the consistency loss on the keyframes predicted by the keyframe predictor and the keyframes predicted by the whole trajectory predictor
\begin{align*}
    L_{cons} = \sum_{i=1}^k ||X_{it} - \tilde{X}_{it}||_2^2.
\end{align*}
In addition, we impose the negative log-likelihood loss on the keyframes
\begin{align*}
    L_{key}(\theta) = - \sum_{j=1}^N I(j=r) 
    \sum_{i=1}^k \log \mathcal{N}(\bar{\mu}_{it} |\mu_{it}, \Sigma_{it};\theta).
\end{align*}
The total loss function is a weighted sum of the losses above
\begin{align*}
    L =L_{traj} + \alpha L_{cons} + \beta L_{key},
\end{align*}
where $\alpha$ and $\beta$ are weights.

\section{EXPERIMENTS}
\subsection{Datasets}
We evaluate our method on two large-scale real world datasets, the Argoverse Forecasting Dataset and the Waymo Open Motion Dataset.
\begin{figure*}[h]
\centering
\includegraphics[width=0.23 \textwidth ]{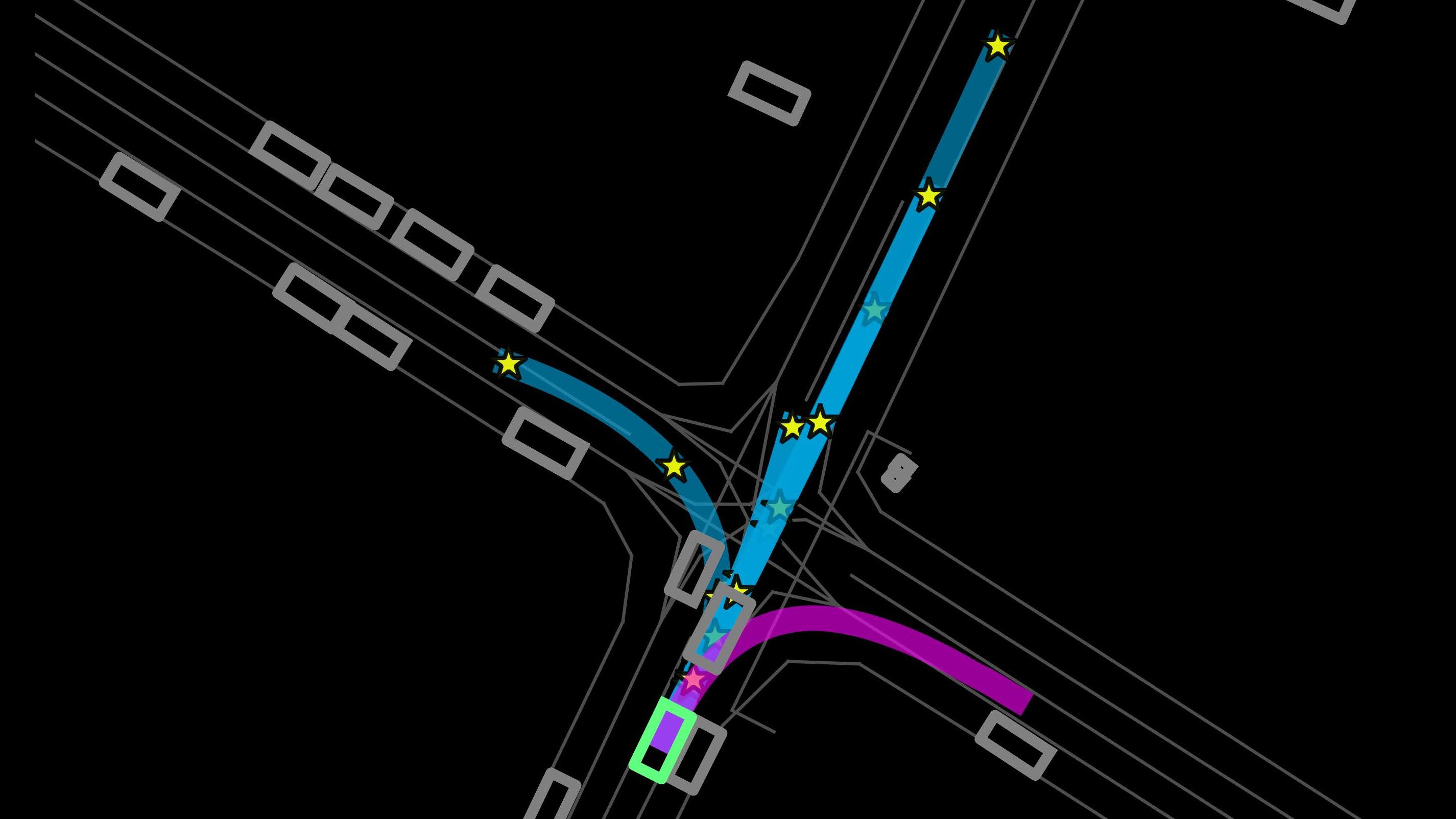}
\includegraphics[width=0.23 \textwidth ]{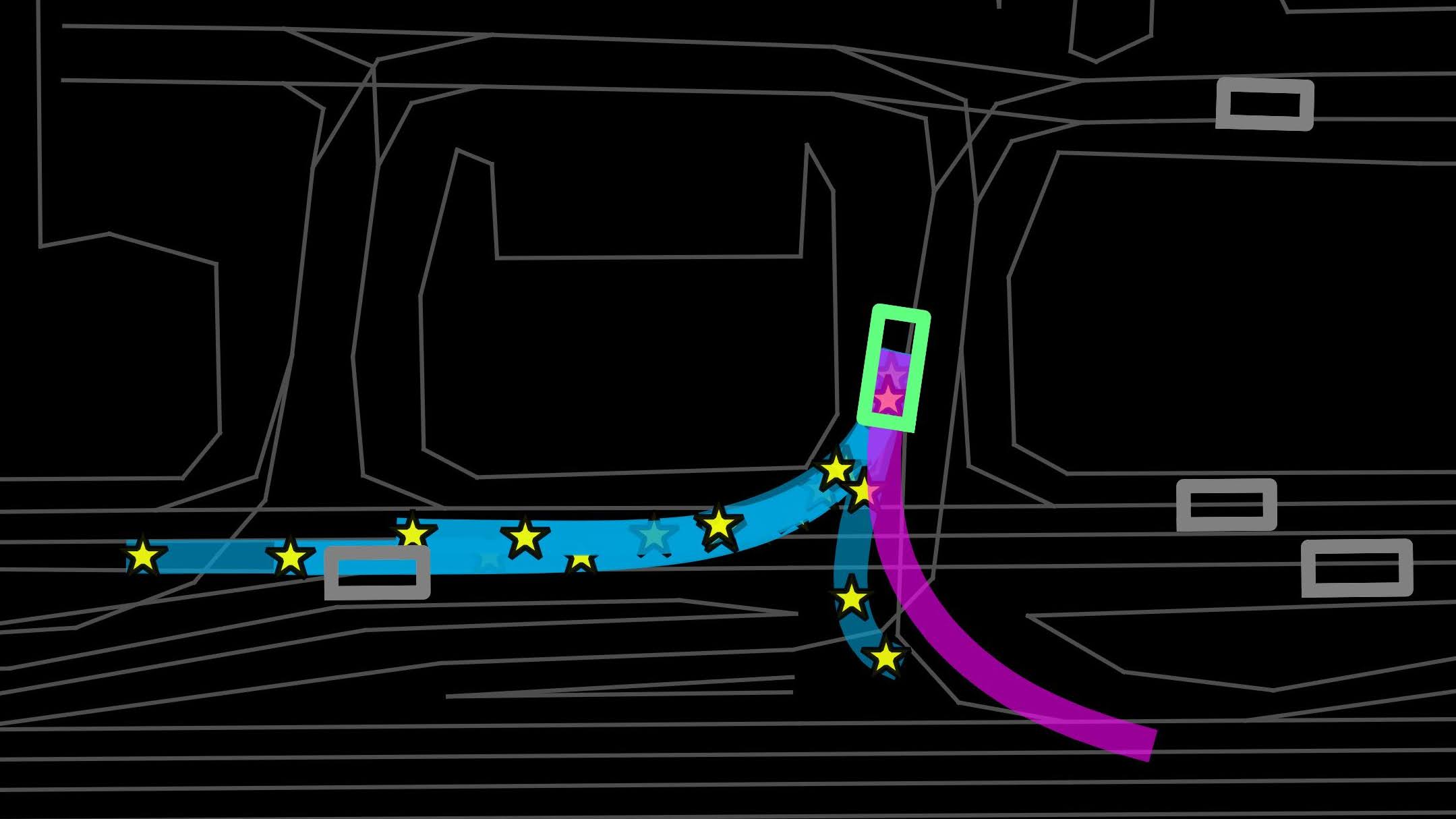}
\includegraphics[width=0.23 \textwidth ]{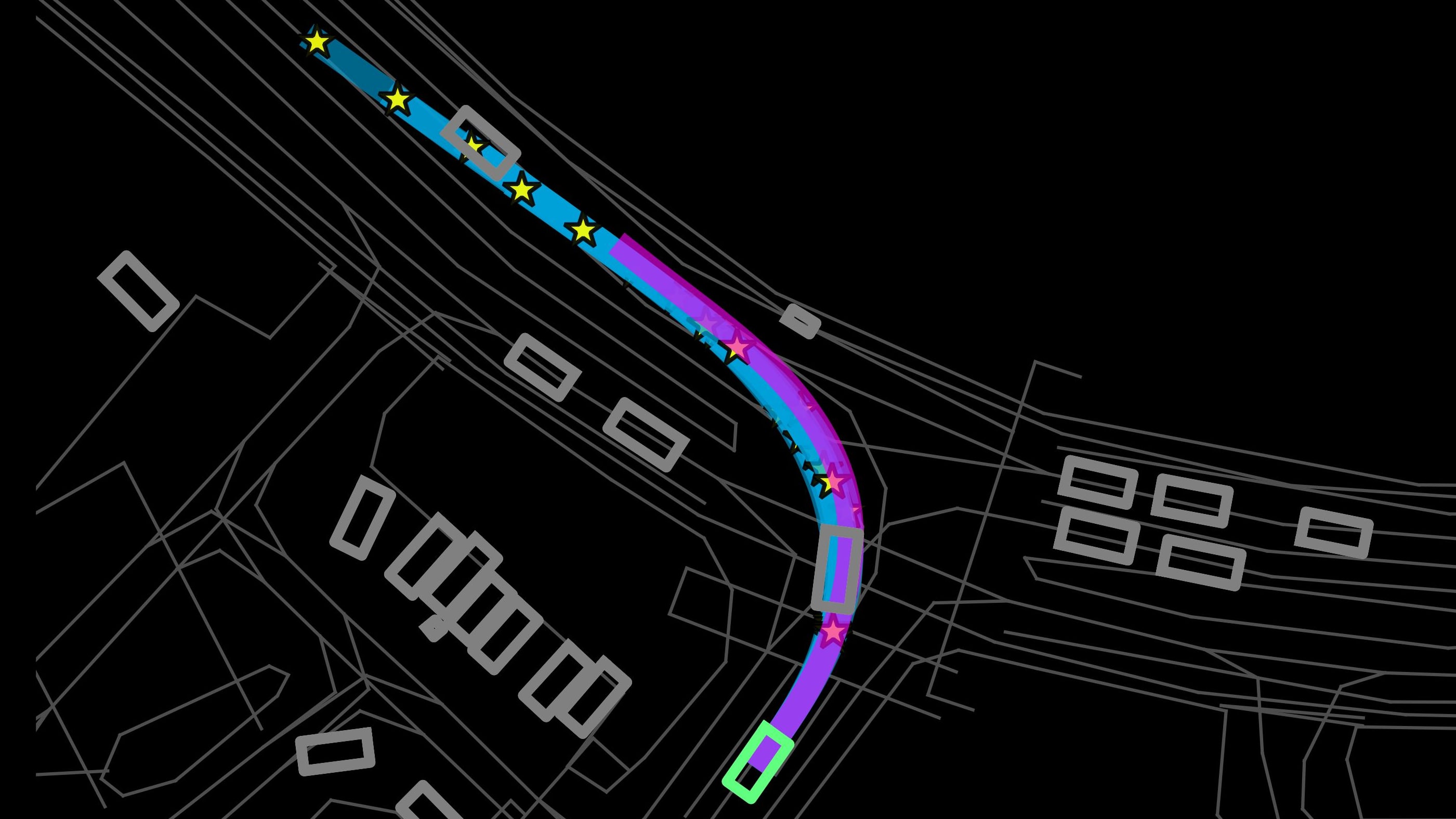}
\includegraphics[width=0.23 \textwidth ]{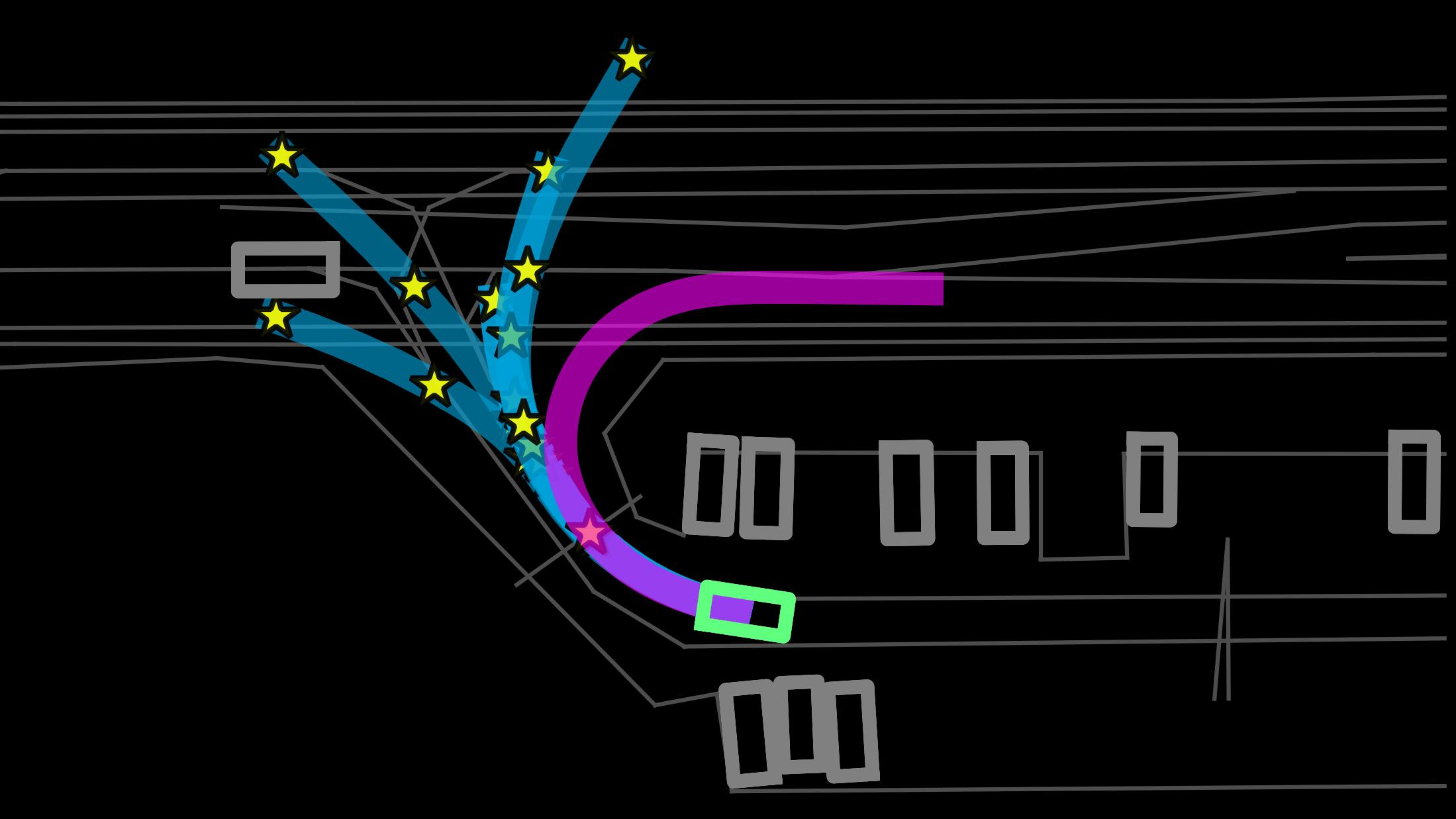}

\vspace{0.05cm}
\includegraphics[width=0.23 \textwidth ]{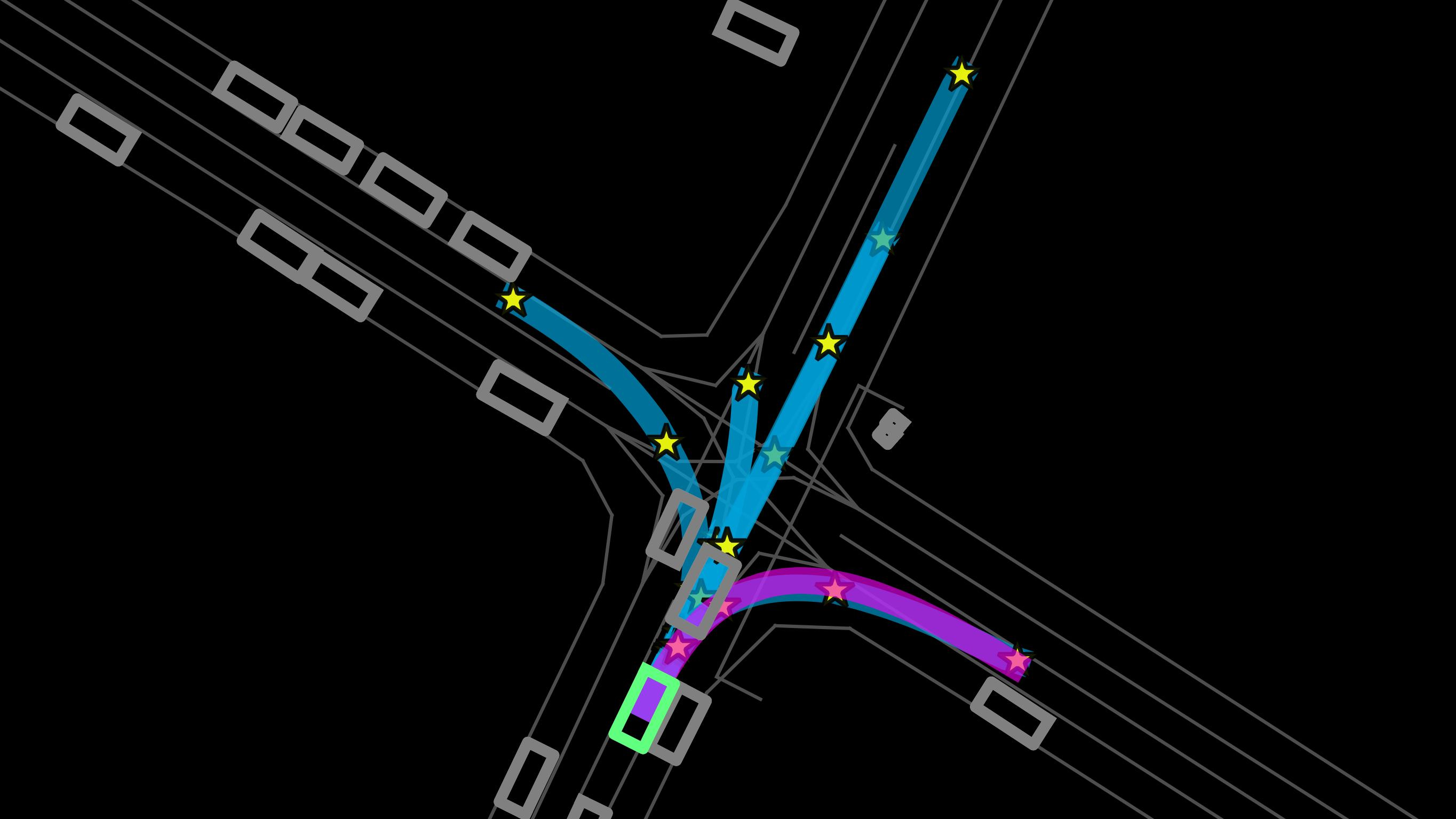}
\includegraphics[width=0.23 \textwidth ]{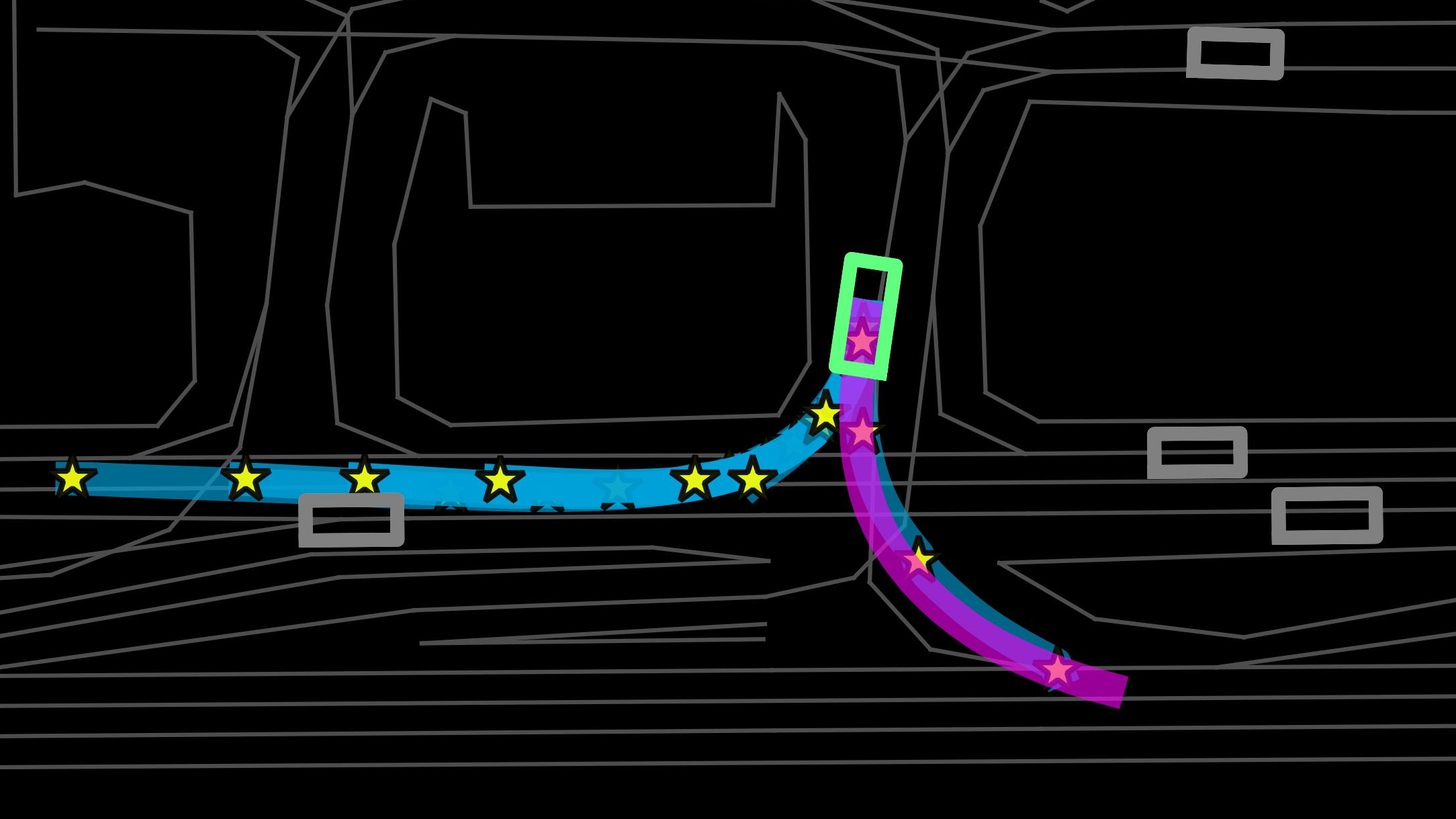}
\includegraphics[width=0.23 \textwidth ]{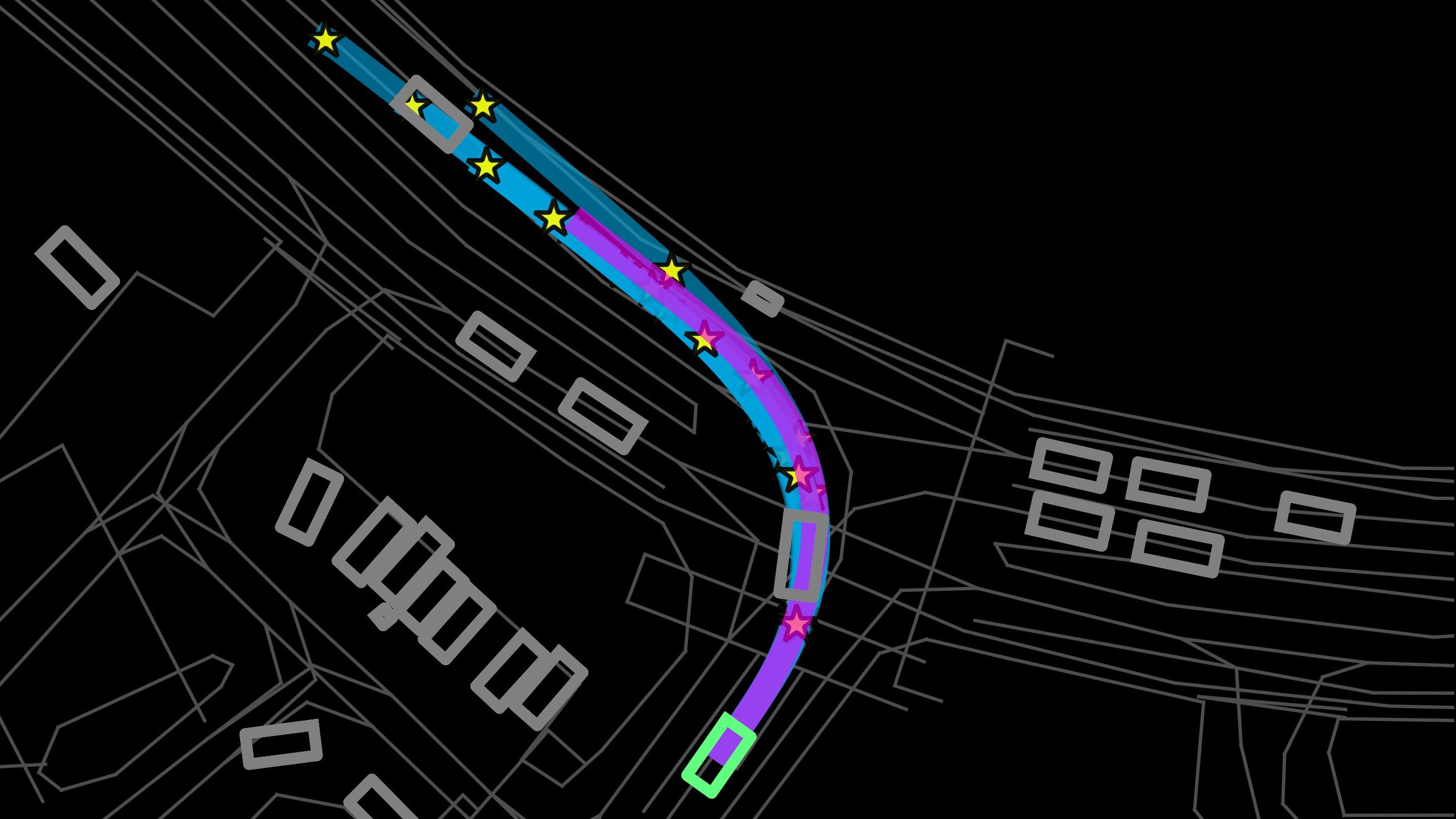}
\includegraphics[width=0.23 \textwidth ]{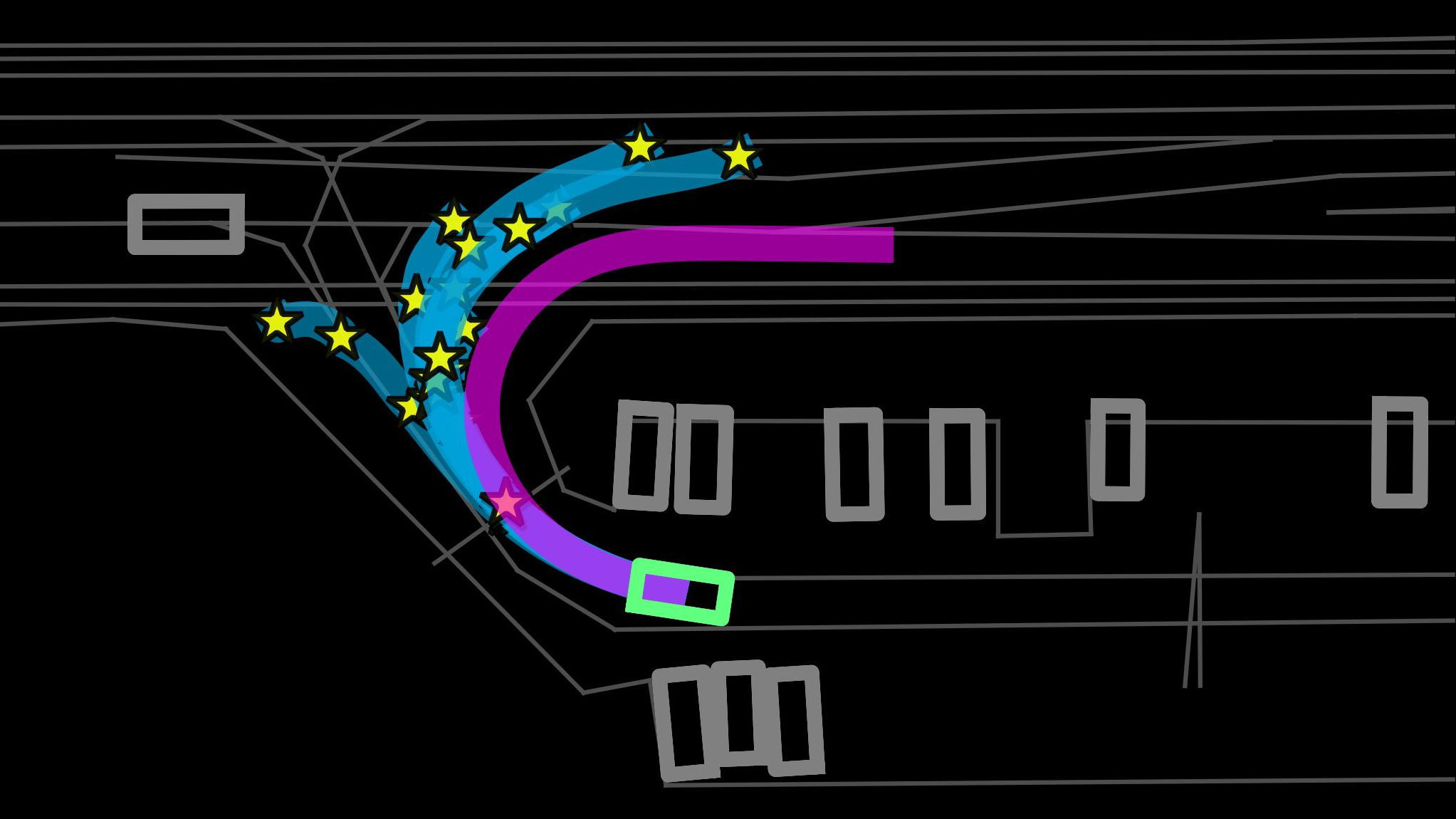}

\vspace{0.05cm}
\includegraphics[width=0.23 \textwidth ]{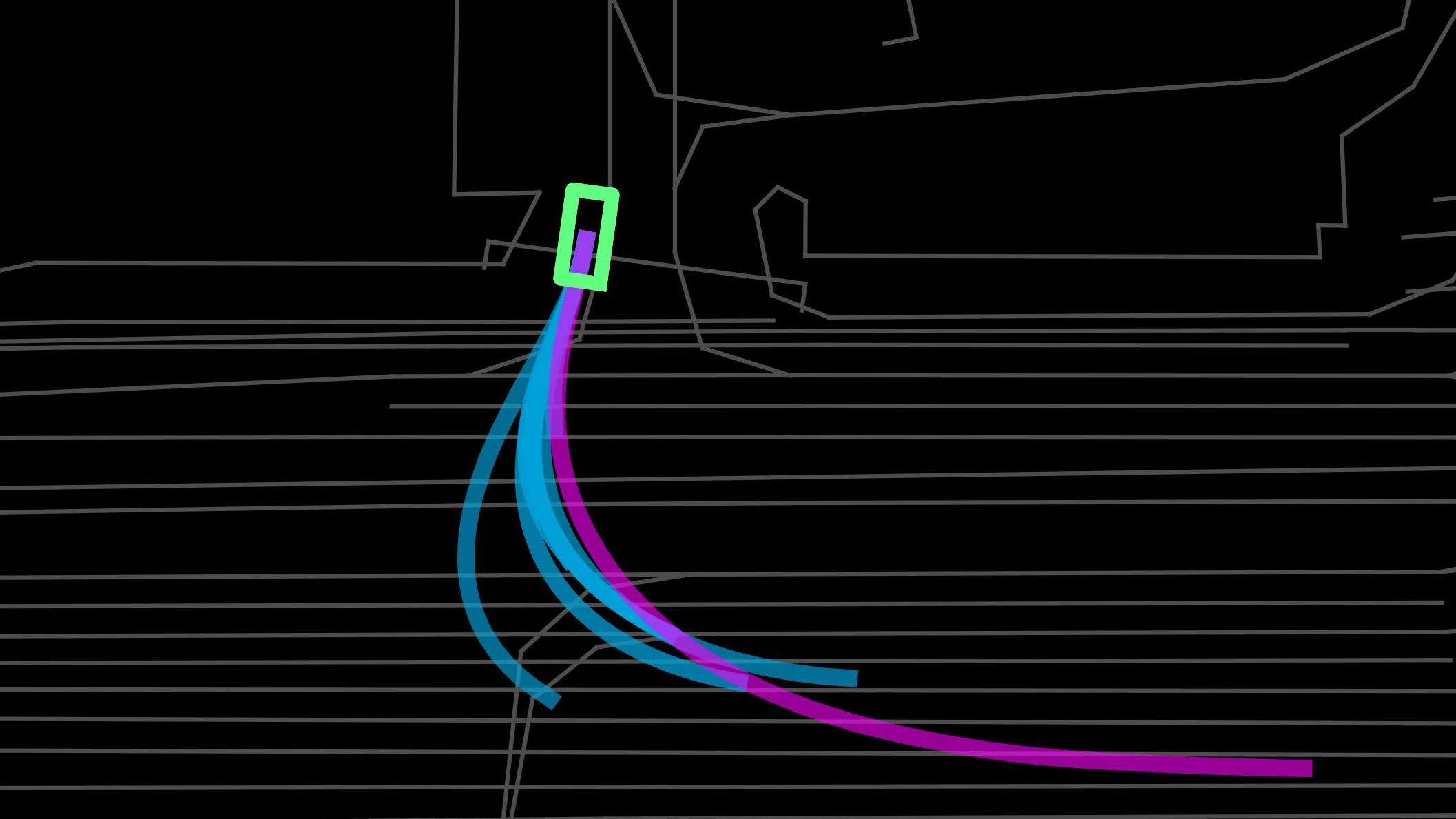}
\includegraphics[width=0.23 \textwidth ]{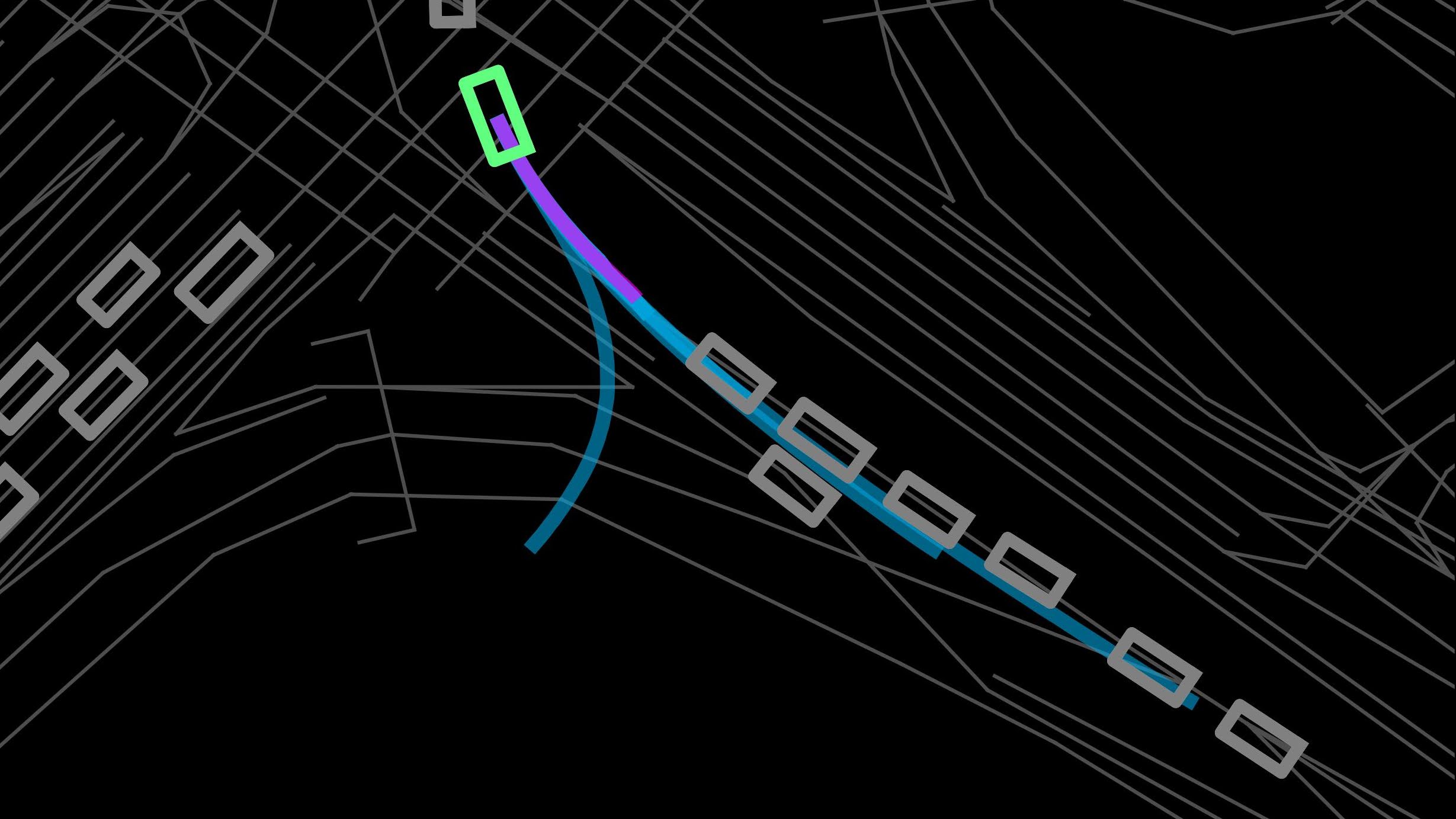}
\includegraphics[width=0.23 \textwidth ]{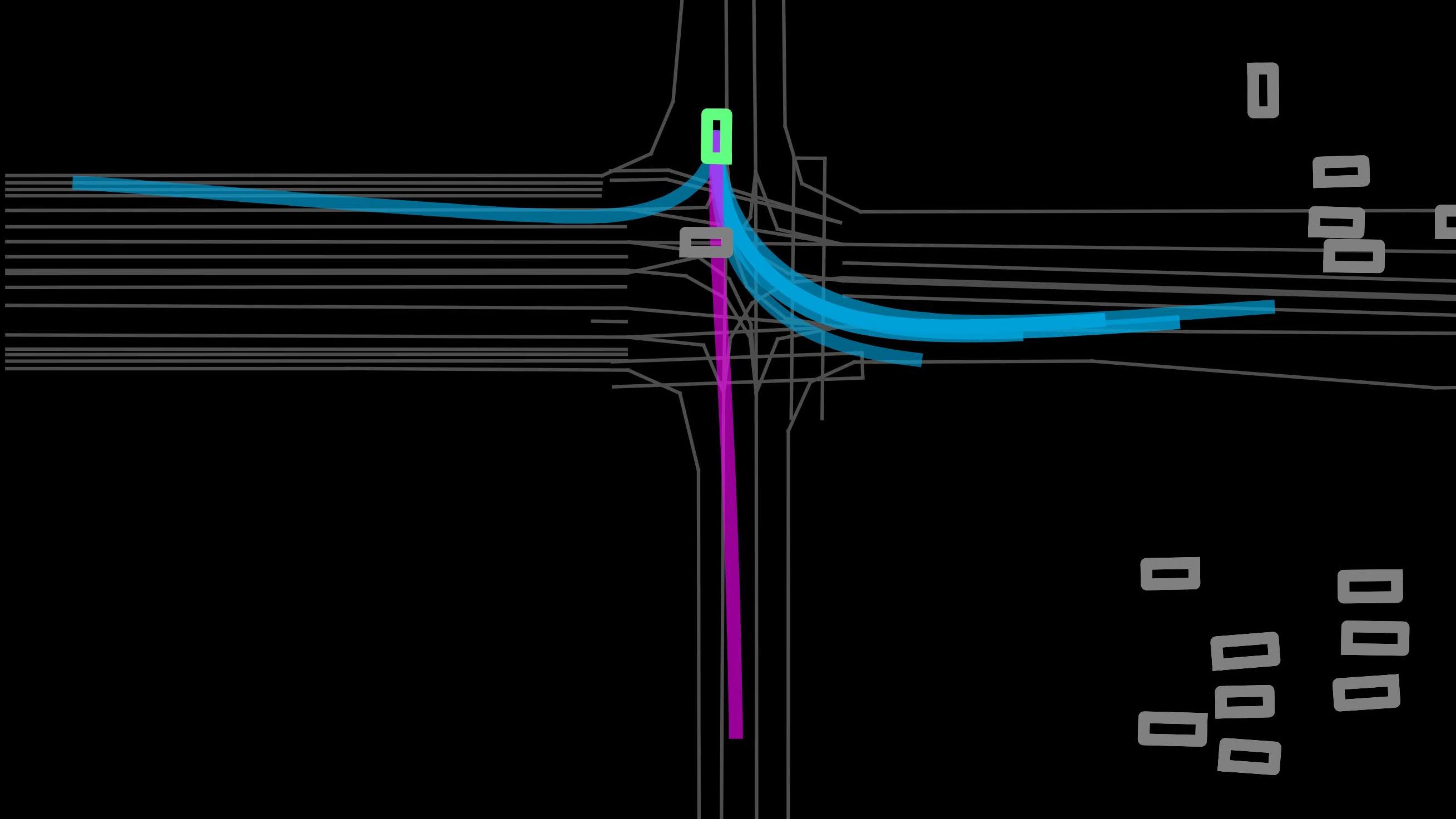}
\includegraphics[width=0.23 \textwidth ]{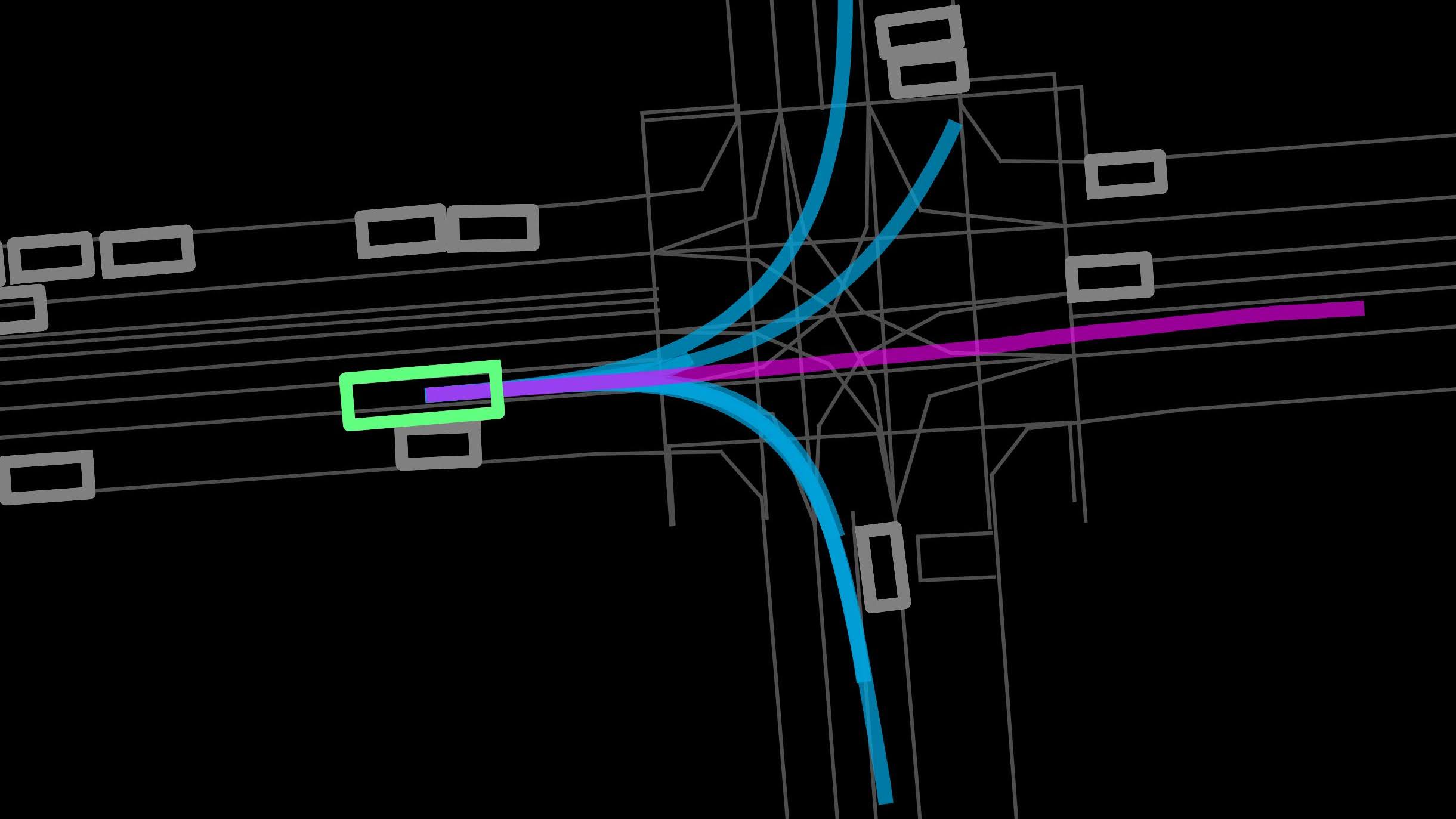}

\vspace{0.05cm}
\includegraphics[width=0.23 \textwidth ]{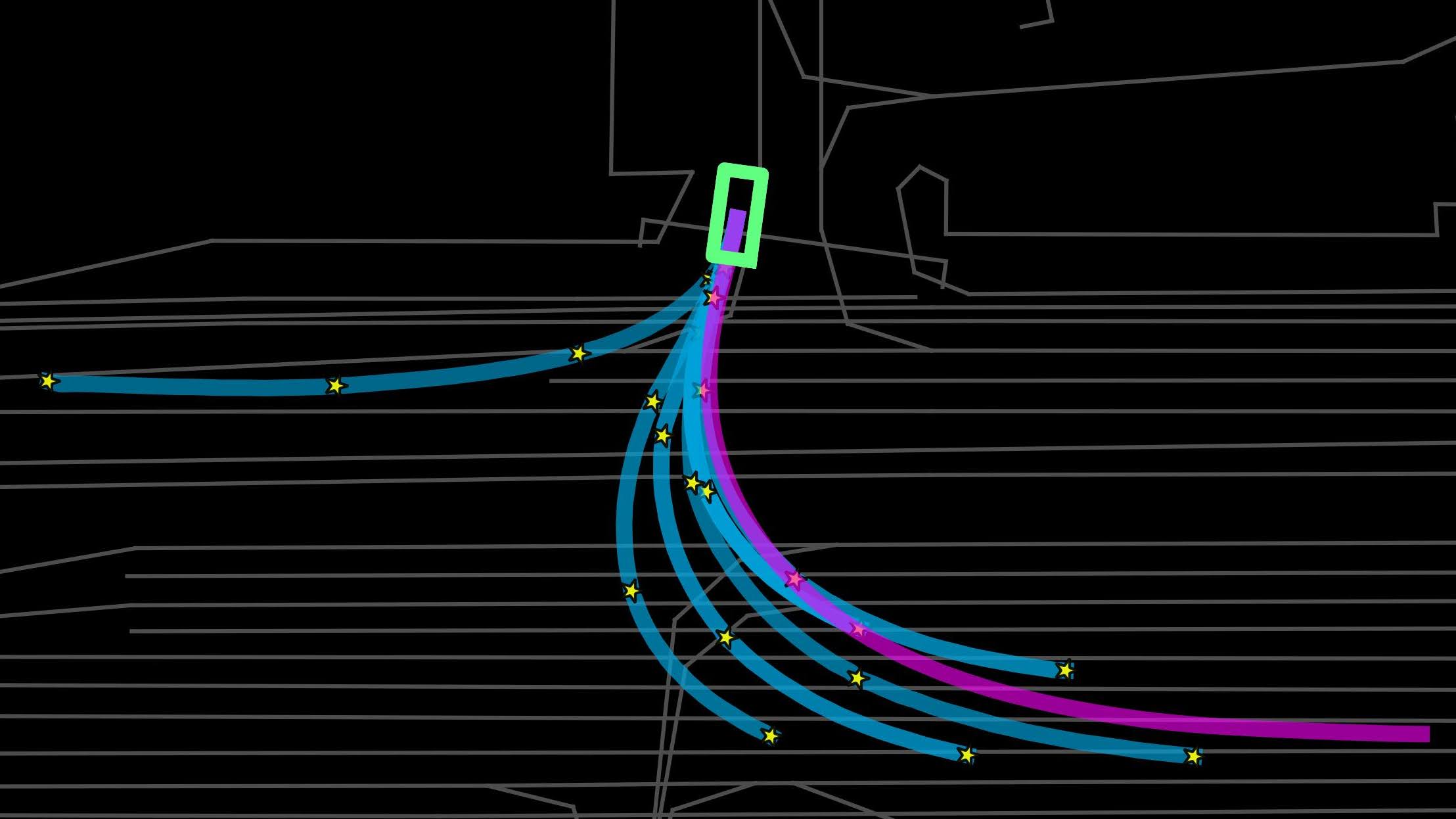}
\includegraphics[width=0.23 \textwidth ]{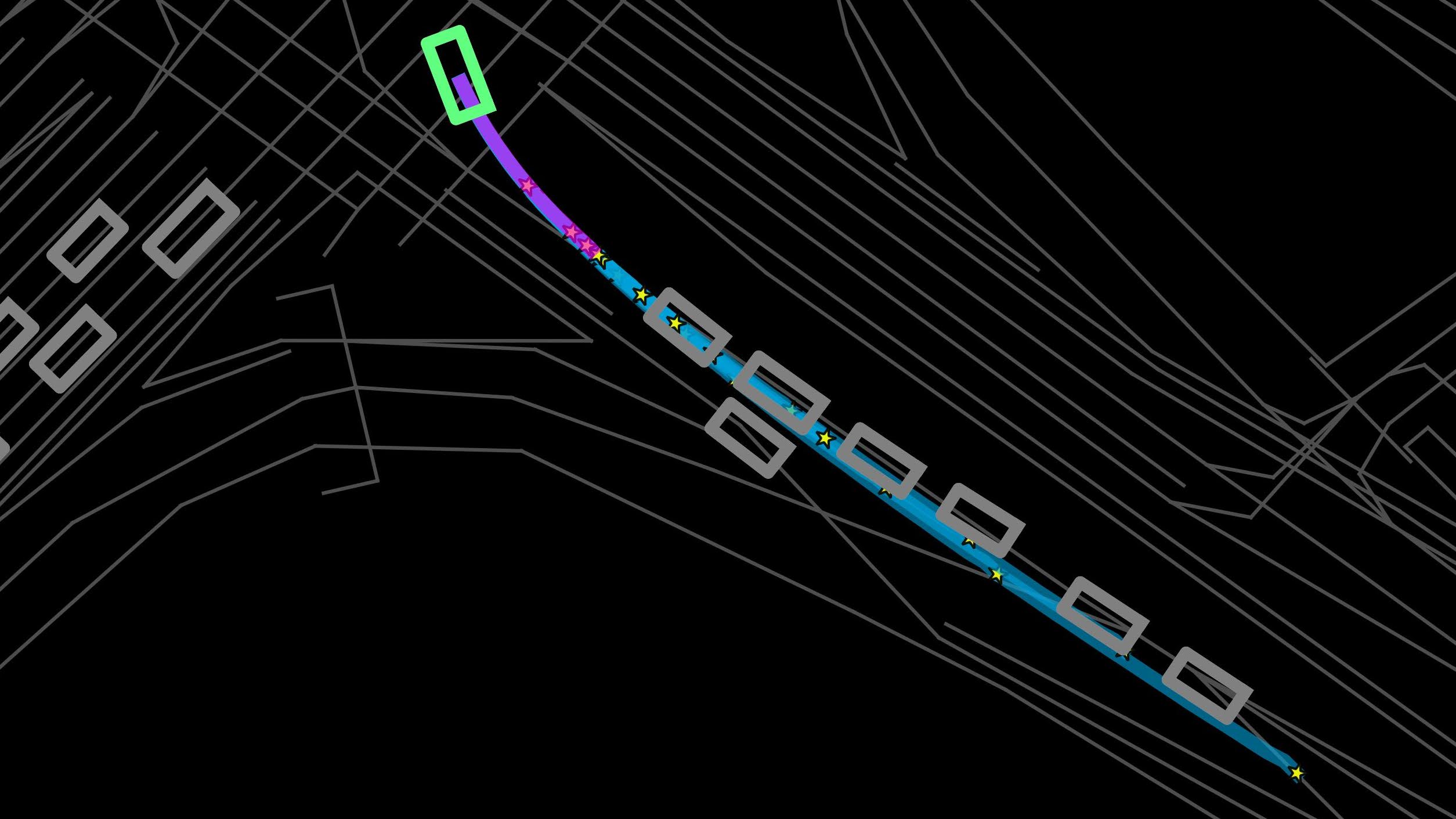}
\includegraphics[width=0.23 \textwidth ]{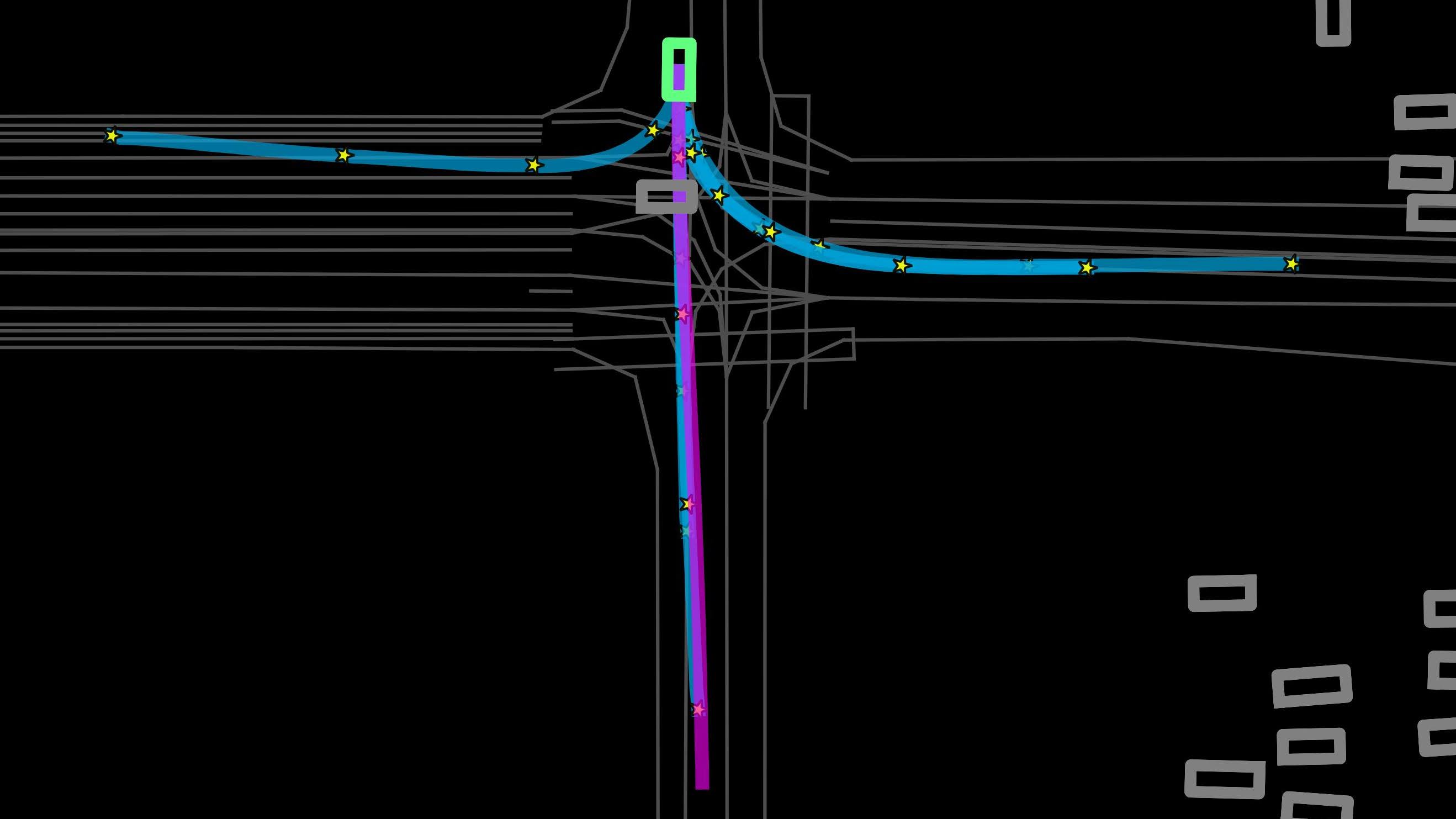}
\includegraphics[width=0.23 \textwidth ]{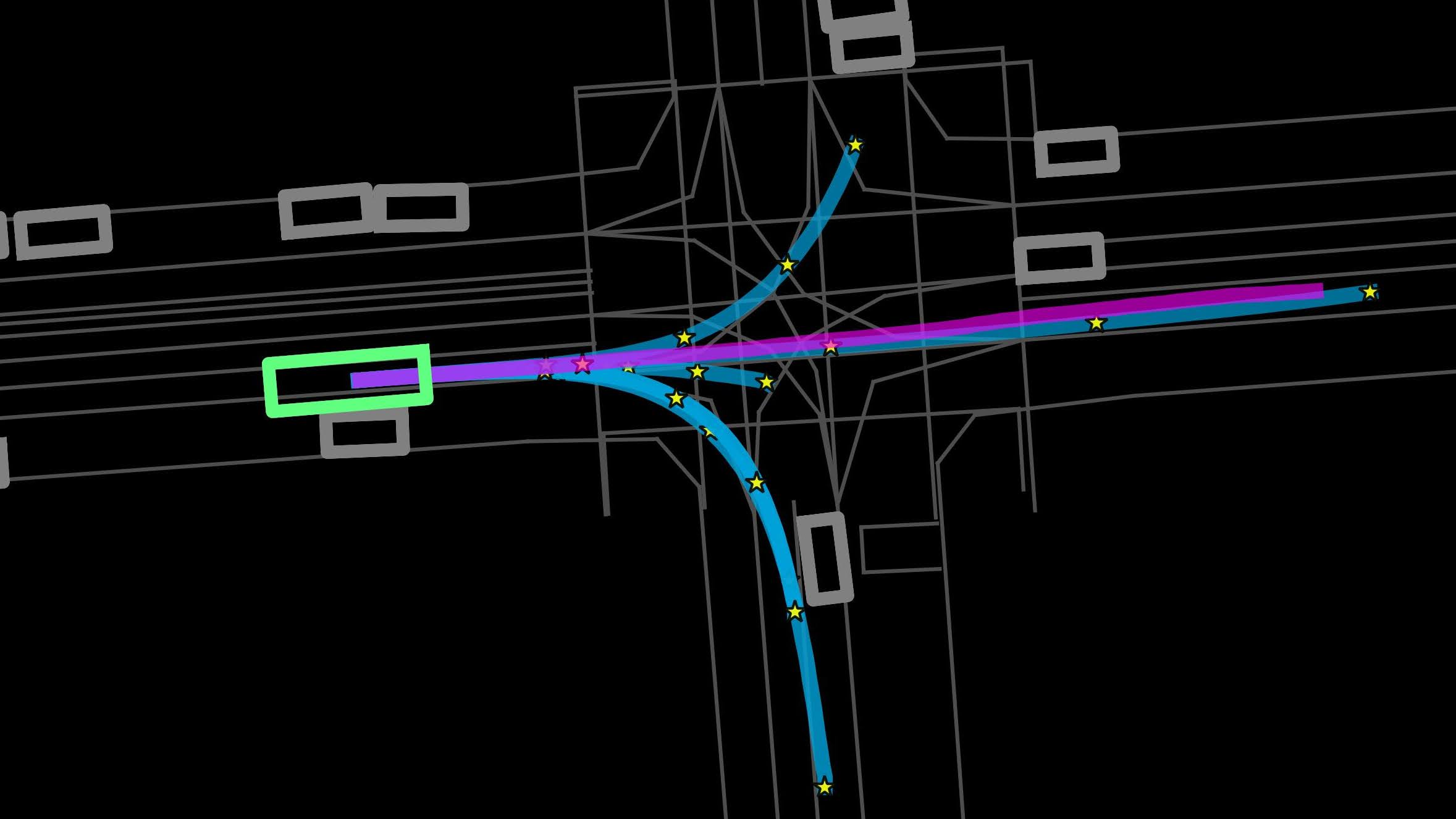}
\caption{\label{fig:example_fig} Samples from WOMD dataset. The agent to be predicted is shown in cyan with its ground truth trajectory shown in magenta. $6$ predicted trajectories are shown in blue with yellow stars annotating the keyframes. We compare KEMP-I-LSTM with Multipath. 1st and 3rd rows: Multipath; 2nd and 4th rows: KEMP-I-LSTM.}
\end{figure*}

\textbf{Argoverse Forecasting Dataset:}
The Argoverse Forecasting Dataset \cite{chang2019argoverse} includes 324,557 five seconds tracked scenarios (2s for the past and 3s for the prediction) collected from 1006 driving hours across both Miami and Pittsburgh. Each motion sequence contains the 2D bird's eye view centroid of each tracked object sampled at 10 Hz. It covers diverse scenarios such as vehicles at intersection, taking turns, changing lanes, and dense traffic. Only one challenging vehicle trajectory is selected as the focus of the forecasting task in each scenario.

\textbf{Waymo Open Motion Dataset:}
The Waymo Open Motion Dataset (WOMD) \cite{sun2020scalability} is by far the largest interactive motion dataset with multiple types of agents: vehicles, pedestrians and cyclists. It consists of 104,000 run segments with over 1,750 km of roadways and 7.64 million unique agent tracks with 20 seconds duration and sampled at 10 Hz. Each segment is further broken into 9 seconds windows (1s for the past and 8 seconds of future data) with 5 seconds overlap. 

\subsection{Metrics}
Given one historical motion, $K$ predictions are output from a model to compare with the ground truth motion. We used both standard metrics and dataset-specific ranking metrics to evaluate our model's performance. $L_2$ distance between a predicted trajectory and the corresponding ground truth is widely used to quantify the displacement error. For the multiple predictions setting, minimum average displacement error (minADE) among all predictions is computed for performance comparison among models. Similarly, minFDE is computed as the minimal $L_2$ distance among the predicted trajectories and ground truth at the last time step (endpoint). Besides these two standard metrics, Miss Rate (MR) is additionally evaluated, which is the number of scenarios where none of the predicted trajectories are within a certain distance of the ground truth according to the endpoint error divided by the total number of predictions. For WOMD, mean Average Precision (mAP) is designed to measure precision-recall performance of the future predictions with a normalized averaged over different types of behavior; we use mAP as the primary model metric.

\subsection{Baseline Algorithms}
As our keyframe-based model can be viewed as a generalization of goal-conditioned models, we contrast its performance with the currently top ranked goal-conditioned model, DenseTNT, as well as other strong baseline models.

\subsection{Implementation Details}
\textbf{Multiple future predictions:}
To obtain more diverse candidate trajectories, our model predicts $m$ trajectories, more than the required number of predictions. To make the predictions more robust, we trained $n$ models independently and ensemble their predictions. At inference time, among all $nm$ candidate trajectories, the top $K$ trajectories are selected using non-maximum suppression algorithm (NMS), where top ranked trajectories (with highest estimated likelihood) are selected greedily while their nearby trajectories are rejected. 

\textbf{Model variants:}
The keyframe predictor and the whole trajectory predictor both output sequences of states. They can either predict the sequence in one shot, or predict the sequence iteratively in an autoregressive fashion. For both cases, we can use an MLP as the predictor (either one-shot or autoregressive), or an LSTM for the autoregressive case. 

\textbf{Training details:}
Our model is trained with a batch size of $256$ on WOMD training dataset and a batch size of $128$ on Argoverse. We set loss weights $\alpha=10, \beta=1$ for all models. Network is trained by ADAM optimizer with learning rate $3\times 10^{-4}$, with an exponential decay of 0.5 every 200k steps for WOMD and 100k for Argoverse. Both models are trained on TPU custom hardware accelerator \cite{jouppi2017datacenter} and converged in $3$ days on WOMD and $2$ days on Argoverse Dataset.

\section{Results}
\begin{table*}[!h]
\caption{Model Performance on Waymo Open Dataset (Leaderboard)}
\label{tab:WOMD}
\begin{center}
\begin{tabular}{|c|c|c|c|c|c|c|c|}
\hline
Model & minADE $\downarrow$ & minFDE$\downarrow$ & MR$\downarrow$ & mAP$\uparrow$ &mAP(3s)$\uparrow$ & mAP(5s)$\uparrow$ & mAP(8s)$\uparrow$\\
\hline
DenseTNT$^{5th}$\cite{gu2021densetnt} & 1.0387 & 1.5514 &0.1779 & 0.3281 & 0.4059 &0.3195 & 0.2589\\
\hline
TVN$^{4th}$ & 0.7498&1.5840 & 0.1833 &0.3341&0.3888&0.3284&0.2852\\
\hline
Scene-Transformer(M+NMS)$^{3rd}$ &0.6784&1.3762 & 0.1977 &0.3370&0.3984&0.3317& 0.2809 \\
\hline
Kraken-NMS$^{2nd}$ & 0.7407	& 1.5786 & 0.2074 &0.3561 & 0.4339&0.3591 & 0.2754\\
\hline
Multipath++$^{1st}$ & 0.5749 & 1.2117 &
0.1475 & 0.3952 &0.4710 & 0.4024& 0.3123\\
\hline
KEMP-I-LSTM (ours) & \textbf{0.5733} & \textbf{1.2088} & \textbf{0.1453} & \textbf{0.3977} &\textbf{0.4729}&\textbf{0.4042}&\textbf{0.3160}\\
\hline
KEMP-I-MLP (ours) & \textbf{0.5723} & \textbf{1.2048} & \textbf{0.1450} & 0.3968 &0.4683&0.4080&0.3141\\
\hline
KEMP-S (ours) & \textbf{0.5714} & \textbf{1.1986} & \textbf{0.1453} & 0.3942 &0.4729&0.4018&0.3080\\
\hline
\end{tabular}
\end{center}
\end{table*}

\begin{table*}[h]
\caption{Ablation study on Waymo Open Dataset (Validation set)}
\label{tab:ab_study1}
\begin{center}
\begin{tabular}{|c|c|c|c|c|c|c|c|}
\hline
Model & minADE$\downarrow$ & minFDE$\downarrow$  & MR $\downarrow$  &\textbf{ mAP }$\uparrow$ &mAP(3s) $\uparrow$& mAP(5s)$\uparrow$ & mAP(8s)$\uparrow$\\
\hline 
KEMP-I-LSTM &0.5718 &1.2061&0.1470&0.3881&0.4735&0.3904&0.3004\\
\hline
KEMP-I-MLP &0.5758&1.2164&0.1487&0.3922&0.4780&0.3995&0.2991\\
\hline
LSTM (No keyframes) & 0.5724&1.2099&0.1482&0.3837&0.4676&0.3879&0.2955\\
\hline
MLP (No keyframes) &0.5736&1.2157&0.1493 &0.3828& 0.4656&0.3892&0.2935\\
\hline
\end{tabular}
\end{center}
\vspace{-0.5cm}
\end{table*}

\begin{table*}[h]
\caption{Ablation study on Waymo Open Dataset (Validation set)}
\label{tab:ab_study2}
\begin{center}
\begin{tabular}{|c|c|c|c|c|c|c|c|}
\hline
Model & minADE$\downarrow$ & minFDE$\downarrow$  & MR $\downarrow$  &\textbf{ mAP }$\uparrow$ &mAP(3s) $\uparrow$& mAP(5s)$\uparrow$ & mAP(8s)$\uparrow$\\
\hline
KEMP-S & {0.5691} & {1.1993} & {0.1458} & 0.3940 &0.4791&0.3959&0.3071\\
\hline
KEMP-S without $L_{cons}$ loss & 0.5698&1.2021&0.1476&0.3949&0.4785&0.4019&0.3043\\
\hline
KEMP-S without $L_{key}$ loss & 0.5710 & 1.2074 & 0.1467 & 0.3955 &0.4783&0.4009&0.3074\\
\hline
KEMP-S without $L_{cons}$ and $L_{key}$ losses &0.5723&1.2103&0.1484 &0.3942& 0.4801&0.4018&0.3008\\
\hline
\end{tabular}
\end{center}
\vspace{-0.5cm}
\end{table*}

\begin{table*}[h]
\caption{Effect of number of keyframes on WOMD (Validation Set)}
\label{tab:ab_study3}
\begin{center}
\begin{tabular}{|c|c|c|c|c|c|c|c|}
\hline
Number of keyframes & minADE$\downarrow$ & minFDE$\downarrow$  & MR $\downarrow$  &\textbf{ mAP }$\uparrow$ &mAP(3s) $\uparrow$& mAP(5s)$\uparrow$ & mAP(8s)$\uparrow$\\
\hline 
0 & 0.5724&1.2099&0.1482&0.3837&0.4676&0.3879&0.2955\\
\hline 
1 &0.5720 &1.2117&0.1466&0.3945&0.4781&\textbf{0.4019}&0.3034\\
\hline
2& \textbf{0.5678} & \textbf{1.1993} & \textbf{0.1454} &0.3881&0.4726&0.3921&0.2995\\
\hline
4& {0.5691} & {1.1993} & {0.1458} & 0.3940 &0.4791&0.3959&0.3071\\
\hline
8& 0.5715&1.2082&0.1490&\textbf{0.3963}&\textbf{0.4800}& 0.3977 & \textbf{0.3110}\\
\hline
16& 0.5735&1.2136&0.1501&0.3894&0.4742&0.3927&0.3012\\
\hline
40 & 0.5737&1.2164&0.1487&0.3915&0.4780&0.3959&0.3006\\
\hline
\end{tabular}
\end{center}
\vspace{-0.5cm}
\end{table*}

\begin{table}[h]
\caption{Model Performance on Argoverse Dataset (Leaderboard)}
\label{tab:Argoverse}
\begin{center}
\begin{tabular}{|c|c|c|c|c|}
\hline
Model & minADE $\downarrow$ & minFDE $\downarrow$& MR $\downarrow$\\
\hline
TNT\cite{zhao2020tnt} &0.94&1.54&13.3\% \\
LaneRCNN\cite{zeng2021lanercnn} & 0.90 & 1.45 & 12.3\% \\
SenseTime\_AP &0.87 &\textbf{1.36}&12.0\% \\
Poly&0.87&1.47&12.0\% \\
PRIME\cite{song2021learning}&1.22&1.56&11.5\% \\
DenseTNT\cite{gu2021densetnt}&0.94& 1.49 & \textbf{10.5}\% \\
\hline
KEMP-I-LSTM & \textbf{0.85} & 1.38 & 12.9\%\\
\hline
\end{tabular}
\end{center}

\vspace{-7mm}

\end{table}


\subsection{Results on benchmarks}
Benchmarks on both datasets are listed in \Cref{tab:WOMD,tab:Argoverse}. 
\textbf{Waymo Open Motion Dataset:} This is a more challenging dataset compared to Argoverse Dataset due to the longer prediction duration and more complex scenarios. 
The models are ranked by mAP.
Our best models are:
\begin{enumerate}
    \item KEMP-I-LSTM in \Cref{tab:WOMD}: An interpolation model, where the keyframe predictor is implemented by LSTM, and the whole trajectory predictor is implemented by MLP. The number of keyframes is 4.
    \item KEMP-I-MLP in \Cref{tab:WOMD}: An interpolation model, where the keyframe predictor and the whole trajectory predictor are implemented by MLP. The number of keyframes is 4.
    \item KEMP-S in \Cref{tab:WOMD}: A separable model, where the keyframe predictor and the whole trajectory predictor are implemented by LSTM. The number of keyframes is 4.
\end{enumerate}

The first five rows in \Cref{tab:WOMD} show the top 5 methods on the Waymo Open Motion Dataset Leaderboard as of September 1st, 2021. KEMP-I-LSTM outperforms baseline models in all metrics.  Additionally, the higher values we achieved in the breakdown of mAP from 3 seconds, 5 seconds and 8 seconds indicate the effectiveness of keyframes as a guidance to the whole trajectory and demonstrate that our model is able to predict trajectories more accurately in the long-term task. 
KEMP-S is better than KEMP-I-LSTM in terms of minADE and minFDE, though it has lower mAP.

\Cref{fig:example_fig} shows qualitative results in different scenarios from WOMD validaton set.
We look at the four examples in the top two rows in detail.
In the first case, both models are able to predict diverse modes (turning left, going straight), while the baseline model fails to predict turning right, which is the agent's actual behavior in the next 8 seconds. In the second case, both models have predicted the correct intent of turning left, but our model has a more natural prediction with keyframes closely aligned to the ground truth. In the third case, our model is able to propose more diverse and reasonable possibilities in the future without missing the mode that agent actually follows. In the fourth case, although both models have diverse predictions spanning the roadgraph space, our model has more reasonable predictions. Compared to the baseline model, KEMP is able to produce more accurate predictions and recall more diverse modes. We believe these good properties are brought by the design of the keyframe architecture. By focusing on the keypoints first to ease the burden of predicting intermediate points, patterns between trajectory and the environment are more easily learned.

\textbf{Argoverse Forecasting Dataset:} 
\Cref{tab:Argoverse} shows several popular methods on Argeverse Dataset Leaderboard, including TNT, DenseTNT, LaneRCNN, and PRIME.
Our model achieves lower minADE and minFDE compared to DenseTNT. In general, we achieve the lowest minADE and second-lowest minFDE among all baseline models, which indicates that KEMP is able to produce realistic trajectories that are very close to the ground truth. However, as the trajectories in Argoverse are fairly short (3 seconds future), our keyframe model does not have a significant advantage over other models.

\subsection{Ablation studies}
First, we compare KEMP against non-keyframe models on the validation set of the Waymo Open Dataset. 
As shown in \Cref{tab:ab_study1}, KEMP-I-MLP has the highest mAP on validation set, though it has also the highest minADE and minFDE.
We also run LSTM and MLP models without keyframe prediction as baselines, and observe that their mAP is more than $1\%$ lower than those of KEMP models. This suggests that keyframes have a positive effect on model quality.

Second, we ablate losses from the KEMP-S model in \Cref{tab:ab_study2}. The fluctuation of mAP among different models is minor: less than $0.3\%$.
The reason might be that the consistency loss and the keyframe loss are complementary given the whole trajectory loss -- getting rid of either does not affect the model much. Even after removing both losses, the keyframe predictor can still learn certain features or latent keyframes, because the whole trajectory predictor predicts segments conditioned on the output of the keyframe predictor. 

Finally, we vary the number of keyframes in the KEMP-S model as shown in \Cref{tab:ab_study3}.
Note that our keyframes are equally spaced. So when the number of keyframes is 1, the model becomes goal-conditioned and hence resembles TNT. 
We observe that 2 keyframes attains the best recall, as the minADE, minFDE, and MR metrics are best. However, the mAP metrics are generally better with 8 keyframes. This indicates that there is a tradeoff that can be made between fewer keyframes, which may increase diversity at the cost of precision, and more keyframes, which provide finer granularity and hence better precision.
With too many keyframes, the model may not be taking advantage of the hierarchical structure of the trajectory prediction problem -- when we go up to 40 keyframes, for example, metrics become worse.
Therefore, depending on whether we care more about precision or recall, the keyframe number can be tuned accordingly.

\section{CONCLUSION AND FUTURE WORK}
In this paper, we proposed a keyframe-based hierarchical end-to-end deep model for long-term trajectory prediction. Our framework generalizes goal-based trajectory prediction methods. Our predictors are automatically learned and does not require hand-crafted algorithms. Our model achieved state-of-the-art performance on the Waymo Open Motion Dataset. 
Future work could try more complicated structure for the keyframe predictor and the whole trajectory predictor for better performance. Another important direction could be a different definition of keyframes. Currently in our model the keyframes are evenly-spaced states. One could try unevenly-spaced states as keyframes. 




\newpage


\bibliographystyle{IEEEtran}
\bibliography{references}

\end{document}